\documentclass[11pt]{article}

\usepackage[final]{acl}

\usepackage{times}
\usepackage{latexsym}

\usepackage[T1]{fontenc}
\usepackage[utf8]{inputenc}

\usepackage{microtype}

\usepackage{inconsolata}

\usepackage{graphicx}

\usepackage{amsfonts}
\usepackage{booktabs}
\usepackage{placeins}

\title{Allocate Before You Embed:\\ Adaptive Visual Input Allocation for Video Embeddings}

\author{
  \textbf{Song Jin}\textsuperscript{1},
  \textbf{Zhongtao Jiang},
  \textbf{Chenglei Shen}\textsuperscript{1},
  \textbf{Huanxuan Liao}\textsuperscript{2}, \\
  \textbf{Haozhe Chi}\textsuperscript{3},
  \textbf{Zhiwei Wang},
  \textbf{Kun Xu},
  \textbf{Yong Liu}\textsuperscript{1}\thanks{\ \ Corresponding author.} \\
  \textsuperscript{1}Gaoling School of Artificial Intelligence, Renmin University of China\\
  \textsuperscript{2}Institute of Automation, Chinese Academy of Science,
  \textsuperscript{3}Peking University\\
  jinsong8@ruc.edu.cn
}

\begin{document}
\maketitle
\begin{abstract}
Large-scale video retrieval requires embedding models to encode long and diverse videos under tight visual-input and inference budgets. Existing methods typically sample a small, fixed set of frames at their original resolution, limiting temporal coverage and ignoring frame importance. Our empirical analysis shows that expanding temporal coverage improves retrieval even under a fixed visual-input budget. Gains are larger when the original per-frame resolution is preserved, highlighting the complementary roles of temporal coverage and spatial fidelity. Motivated by this finding, we propose AllocEmbed, an allocate-then-embed framework that reallocates a fixed visual-input budget across more frames. A lightweight allocator uses low-cost previews to assign frame-wise resolutions before the embedding backbone, preserving more detail where it most benefits retrieval while reducing visual cost elsewhere. We further introduce Retrieval-Driven Policy Optimization (RDPO), which learns the allocator directly from retrieval feedback using a rank-validated similarity gap and a confidence-guided efficiency incentive. Operating entirely before the backbone, AllocEmbed integrates with existing retrieval systems without modifying the embedding model or downstream pipeline. Experiments on the MMEB-V2 V-QA and V-RET tasks and our LongRet benchmark show that AllocEmbed achieves the best overall retrieval performance among the evaluated budget-matched methods and transfers across embedding backbones. Our code is publicly available at \url{https://github.com/jinsong8/AllocEmbed}.
\end{abstract}

\section{Introduction}

Video embeddings provide a common interface for large-scale video retrieval, cross-modal search, and retrieval-augmented video understanding. By compressing a variable-length video into a fixed-dimensional vector, they make it possible to compare video content efficiently at corpus scale. In practice, however, video embedding models operate under limited visual context, and running large models over massive video collections is expensive. Existing systems therefore commonly control cost by uniformly sampling a small, fixed number of frames from each video~\citep{luo2022clip4clip,ma2022x,meng2025vlm2vec}. This strategy is simple and cost-effective, but it also assumes that different videos, time positions, and visual evidence carry the same value for retrieval.

\begin{figure}[t]
\centering
\includegraphics[width=\columnwidth]{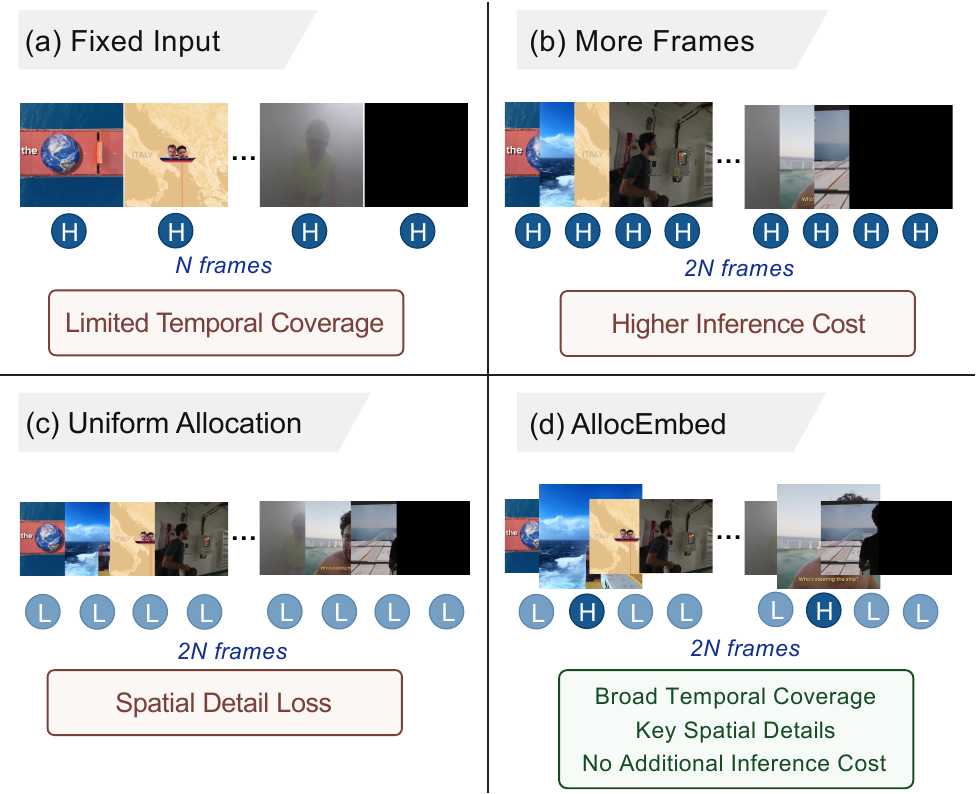}
\caption{Fixed-frame inputs limit temporal coverage. Adding full-resolution frames increases inference cost, whereas uniform downscaling loses spatial detail. AllocEmbed preserves broad temporal coverage and key spatial detail under the same visual-input budget. H/L denote high/low resolution.}
\label{fig:motivation}
\end{figure}

Figure~\ref{fig:motivation} illustrates our motivation. We first examine how temporal coverage affects video retrieval. Across four embedding backbones, sampling more frames improves retrieval when per-frame resolution is preserved. Under a fixed visual-input budget, denser sampling remains beneficial despite the resulting lower per-frame resolution, but yields smaller gains. Together, these results show that temporal coverage and spatial fidelity provide complementary benefits and motivate a central question: how should a fixed visual-input budget be allocated across a denser frame sequence?

We therefore propose AllocEmbed, an allocate-then-embed framework that redistributes a target visual-input budget across a denser frame sequence. Given low-cost previews of the sampled frames and the task text, a lightweight allocator predicts a continuous resolution scale for each frame and constructs a mixed-resolution video input. All sampled frames remain in their original temporal order, while their resolutions are adjusted to preserve more detail where it most benefits retrieval and reduce visual cost elsewhere. The allocated input is then processed by a frozen embedding backbone. Because allocation occurs entirely before embedding, AllocEmbed requires no changes to the backbone weights, embedding dimensionality, similarity function, or downstream retrieval pipeline.

Learning such an allocator is challenging because existing datasets provide no frame-level supervision for resolution, and an allocation can be evaluated only after preprocessing, embedding, and retrieval. We address this problem with Retrieval-Driven Policy Optimization (RDPO), which learns directly from end-to-end retrieval outcomes. RDPO samples multiple allocations for the same input and evaluates them against an augmented candidate set with in-batch and global negatives. It then constructs a group-relative learning signal from a rank-validated similarity gap, which grades only allocations that rank the positive target first. A success-gated, confidence-guided efficiency incentive further favors lower-cost allocations without rewarding cheap but incorrect decisions. This formulation trains the allocator through the actual visual preprocessing and tokenization pipeline while keeping the embedding backbone frozen.

We evaluate AllocEmbed on ten MMEB-V2 video QA and retrieval tasks and on LongRet, a new benchmark for long videos with sparse and unevenly distributed evidence. The allocator is trained with VLM2Vec-V2-2B and evaluated without retraining on three additional embedding backbones, covering model sizes from 2B to 8B parameters. Averaged across the ten MMEB-V2 tasks, AllocEmbed improves Hit@1 by 1.34--1.69 points over the corresponding fixed-frame baselines. Averaged across the four LongRet subsets, the gains increase to 5.69--9.09 points. These aggregate improvements are achieved without exceeding the visual-input cost of each backbone's fixed-frame baseline.

Our contributions are threefold:
\begin{itemize}
    \item We introduce AllocEmbed, an allocate-then-embed framework motivated by our finding that temporal coverage and spatial fidelity provide complementary benefits for video retrieval. To our knowledge, AllocEmbed is the first framework to study visual input allocation for VLM-based video embedding models.
    \item We develop RDPO, a retrieval-driven policy optimization method that learns the allocation policy directly from ranking feedback using a rank-validated similarity gap and a confidence-guided efficiency incentive.
    \item We conduct extensive experiments on MMEB-V2 and our newly constructed LongRet benchmark. AllocEmbed improves aggregate retrieval performance under matched visual-input budgets and transfers from its training backbone to three additional embedding backbones without retraining.
\end{itemize}

\begin{figure*}[!t]
\centering
\includegraphics[width=\textwidth]{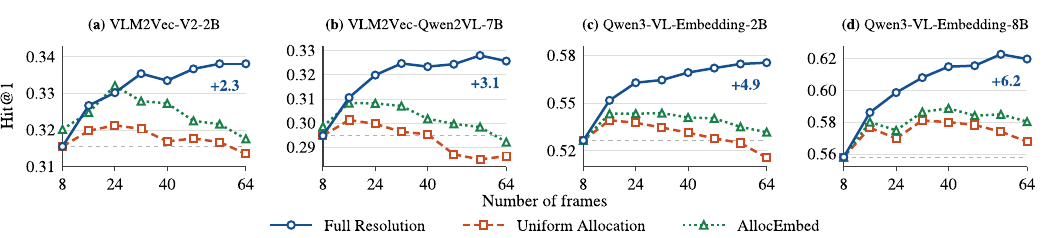}
\caption{Effect of increasing the sampled frame count on mean Hit@1 across ten MMEB-V2 V-QA and V-RET tasks. Full Resolution keeps the original per-frame resolution as the frame count increases, whereas Uniform Allocation and AllocEmbed use the same visual-input budget as the 8-frame input. Blue endpoint labels show the Full Resolution gains relative to the 8-frame baseline. Each panel uses a model-specific vertical range.}
\label{fig:when-more-frames-help}
\end{figure*}

\section{Related Work}

\paragraph{Video and Multimodal Embedding Models.}
Video embedding models enable cross-modal retrieval by mapping videos and natural-language descriptions into a shared representation space. Early approaches either pretrain dedicated video--text encoders, such as VideoCLIP and Frozen in Time~\citep{xu2021videoclip,bain2022frozen}, or adapt image--text models to video through frame-level aggregation, as in CLIP4Clip and X-CLIP~\citep{luo2022clip4clip,ma2022x}. Recent VLM-based embedding models broaden this scope to heterogeneous queries and targets. VLM2Vec supports diverse image--text tasks, while GME unifies retrieval across text, images, composed image--text inputs, and visual documents~\citep{jiang2024vlm2vec,zhang2024gme}. VLM2Vec-V2 further incorporates videos and visual documents and introduces the MMEB-V2 benchmark~\citep{meng2025vlm2vec}, while Qwen3-VL-Embedding encodes heterogeneous inputs, including videos, in a single model~\citep{li2026qwen3}. For video retrieval, however, inputs are still typically constructed from a fixed number of frames at a uniform resolution. The trade-off between temporal coverage and per-frame spatial detail under a constrained visual-input budget therefore remains underexplored.

\paragraph{Efficient Video Processing and Budget Allocation.}
Efficient video processing typically reduces either the number of input frames or the visual representations processed by the model. Frame-selection methods retain sparse salient or query-relevant frames~\citep{buch2025flexible}, while token merging and pruning compress visual representations within VLMs~\citep{ryoo2024xgen,shen2024longvu,shen2026fastvid}. Other work adapts input resolution to video content for efficient action recognition and query-aware video understanding~\citep{meng2020ar,zhang2025q}. Most closely related, ResAdapt learns input-side resolution assignments from final task outcomes without frame-level resolution labels while keeping the backbone unchanged~\citep{liao2026resadapt}. Adaptive-resolution methods generally target classification, answer generation, or reasoning. Efficient video retrieval methods instead rely mainly on CLIP-style encoders and reduce cost through frame selection or frame-level aggregation~\citep{wu2023empirical,hu2023adaclip,shen2025tempme}. VLM-based embedding models differ because multiple frames share the visual context, and frame count and per-frame resolution jointly determine token cost. AllocEmbed targets this setting by preserving a denser sampled sequence and learning per-frame input resolutions directly from retrieval ranking feedback without modifying the backbone.

\begin{figure*}[t]
\centering
\includegraphics[width=0.97\textwidth]{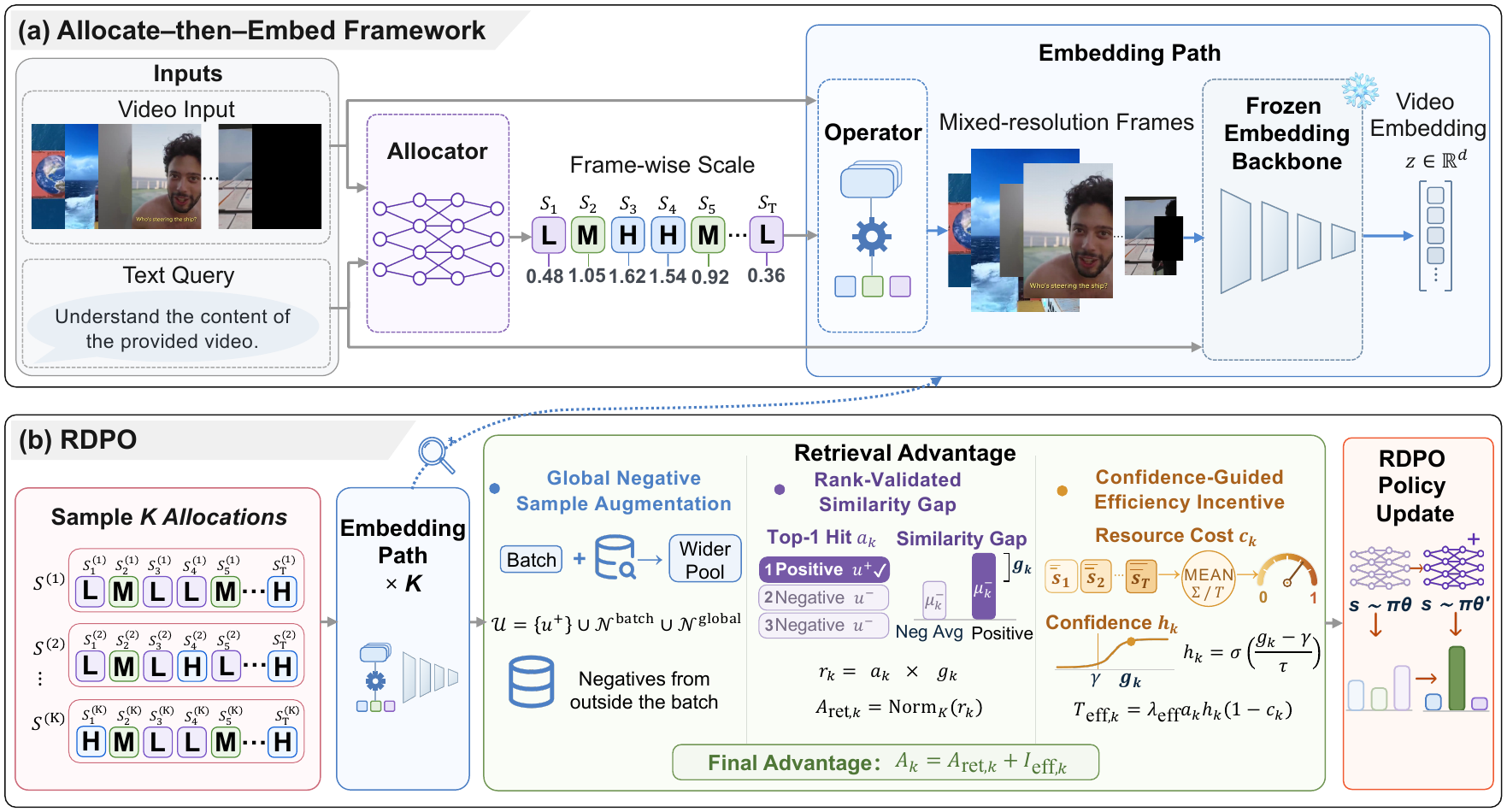}
\caption{Overview of AllocEmbed. (a) The allocator predicts per-frame resolution scales from the video and task text, and a scale-aware operator constructs a mixed-resolution input for the frozen embedding backbone. (b) RDPO evaluates multiple allocations against a candidate set augmented with global negatives, then updates the allocator using a group-relative retrieval advantage based on the rank-validated similarity gap and a confidence-guided efficiency incentive.}
\label{fig:allocembed-overview}
\end{figure*}

\section{Temporal Coverage and Spatial Fidelity}
\label{sec:more-frames}

To study the interplay between temporal coverage and spatial fidelity in video retrieval, we vary the sampled frame count under two visual-input budget regimes. Full Resolution preserves the original per-frame resolution, allowing the visual input to grow with the number of frames. Under the fixed-budget regime, Uniform Allocation accommodates additional frames by resizing them uniformly to an 8-frame-equivalent budget. Within each comparison, the embedding backbone, dataset, evaluation metric, and temporal sampling rule remain unchanged. Figure~\ref{fig:when-more-frames-help} reports mean Hit@1 across ten MMEB-V2 video question answering (V-QA) and video retrieval (V-RET) tasks for four backbones spanning different model families and sizes.

\paragraph{Observation 1: More frames help when per-frame resolution is preserved.}
Across all four backbones, denser sampling yields clear overall gains when the original per-frame resolution is preserved. This consistent trend indicates that conventional fixed-frame inputs leave useful temporal evidence unobserved. Sampling more frames increases the chance of capturing brief or unevenly distributed events, while preserving their resolution keeps the corresponding visual cues accessible to the embedding backbone.

\paragraph{Observation 2: More frames also help under a fixed visual budget.}
Under a fixed visual-input budget, moderately increasing the number of uniformly resized frames improves retrieval across all four backbones. The added temporal evidence can therefore compensate for the accompanying reduction in per-frame resolution. This benefit eventually diminishes as further increases force more aggressive downscaling and the resulting loss of spatial detail begins to outweigh the gain from broader temporal coverage.

Together, these findings show that temporal coverage and spatial fidelity must be balanced under a fixed visual-input budget. AllocEmbed therefore retains a denser frame sequence while allocating resolution non-uniformly, avoiding the uniform loss of spatial detail imposed by global downscaling.

\section{AllocEmbed}
\label{sec:allocembed}

\subsection{Problem Formulation}
\label{sec:problem-formulation}

Let \(V\) denote a video and \(q\) the task-related text input, such as a retrieval instruction or question. The embedding model must preserve retrieval-relevant visual evidence under a limited visual-input budget. Standard pipelines control cost by sampling a small, fixed number of frames at a common resolution, but their restricted temporal coverage can miss brief or sparsely distributed evidence.

We instead sample a denser candidate sequence,
\begin{equation}
X=(x_1,\ldots,x_T),
\label{eq:candidate-sequence}
\end{equation}
where \(T\) exceeds the standard frame count. Because the budget prevents all \(T\) frames from being encoded at high resolution, we introduce a frame-wise allocation vector
\begin{equation}
\mathbf{s}=(s_1,\ldots,s_T),
\label{eq:allocation-vector}
\end{equation}
where \(s_t\) controls the spatial information retained from \(x_t\). Let \(U(\mathbf{s})\) and \(C(\mathbf{s})\) denote the resulting retrieval utility and visual-input cost, respectively. For a budget \(B\), the desired allocation is
\begin{equation}
\mathbf{s}^{\star}
=\mathop{\mathrm{arg\,max}}_{\mathbf{s}:C(\mathbf{s})\leq B}
U(\mathbf{s}).
\label{eq:budgeted-allocation}
\end{equation}
AllocEmbed learns a content- and text-conditioned allocation that redistributes resources across the sampled frames, expanding temporal coverage without uniformly sacrificing the spatial detail required for accurate ranking.

\subsection{Allocate-Then-Embed Framework}
\label{sec:framework}

As illustrated in Figure~\ref{fig:allocembed-overview}(a), AllocEmbed follows an allocate-then-embed design. Given a densely sampled candidate sequence \(X\) and task text \(q\), it first uses a lightweight allocator to predict a spatial scale for each candidate frame:
\begin{equation}
\mathbf{s}=\pi_{\theta}(X,q)=(s_1,\ldots,s_T).
\label{eq:allocator-policy}
\end{equation}
Each \(s_t\) determines the input resolution of frame \(x_t\). The scale-aware operator then resizes the frames according to their assigned scales, producing a mixed-resolution sequence. This sequence and \(q\) are passed to the video embedding backbone to obtain the final embedding. Because allocation and scaling take place before embedding, AllocEmbed can be integrated with existing video embedding models without modifying their backbone architectures, parameters, or downstream retrieval pipelines. It improves retrieval through more effective use of the same visual-input budget.

\subsubsection{Allocator Architecture}
Our allocator implementation is based on ResAdapt~\citep{liao2026resadapt}. A frozen SmolVLM2-256M feature extractor~\citep{marafioti2025smolvlm} first obtains patch-level visual features from low-cost frame previews \(P(X)\) and token-level text features from the task text \(q\). A lightweight frame-representation module then projects the two streams into a shared space, performs cross-modal fusion and spatiotemporal refinement, and aggregates the refined patch features into one representation per frame:
\begin{equation}
(h_1,\ldots,h_T)
=G_{\theta}\!\left(F_{\mathrm{Smol}}(P(X),q)\right).
\label{eq:allocator-features}
\end{equation}
Here, \(F_{\mathrm{Smol}}\) and \(G_{\theta}\) denote the frozen SmolVLM feature extractor and the trainable frame-representation module, respectively. Each \(h_t\in\mathbb{R}^{d}\) represents candidate frame \(x_t\) through its visual content, task relevance, and context within the candidate sequence. A lightweight policy head maps each \(h_t\) to the parameters of a frame-specific scale distribution, as described in the next subsection. Throughout training, \(F_{\mathrm{Smol}}\) remains frozen, and only \(G_{\theta}\) and the policy head are optimized.

\subsubsection{Scale Policy Parameterization}
For each candidate frame, the policy head maps \(h_t\) to the positive shape parameters of a Beta distribution:
\begin{equation}
(\alpha_t,\beta_t)=\phi\!\left(H_{\theta}(h_t)\right),
\quad \alpha_t>0,\ \beta_t>0.
\label{eq:beta-parameters}
\end{equation}
Here, \(H_{\theta}\) is a lightweight residual MLP, and \(\phi\) applies Softplus~\citep{dugas2000incorporating} element-wise to ensure that both parameters are positive.
A normalized action \(a_t\) is drawn from the predicted Beta distribution and linearly mapped to the allowed scale range:
\begin{equation}
\scalebox{0.9}{$\displaystyle
a_t\sim\mathrm{Beta}(\alpha_t,\beta_t),
\quad
s_t=s_{\min}+a_t(s_{\max}-s_{\min}).
$}
\label{eq:scale-sampling}
\end{equation}
Since the Beta distribution is supported on \([0,1]\), the mapped scale satisfies \(s_t\in[s_{\min},s_{\max}]\). By default, we retain the training-time scale range at inference. The learned policy then assigns \(s_t<1\) to most frames, allowing the backbone to process more frames within the same visual budget. Additionally, adjusting \(s_{\min}\) and \(s_{\max}\) can provide further control over the visual budget without retraining.

\subsubsection{Scale-Aware Preprocessing}
For a candidate frame \(x_t\) with spatial size \(H_t\times W_t\), the predicted scale \(s_t\) gives:
\begin{equation}
H'_t\approx s_tH_t,\quad
W'_t\approx s_tW_t,\quad
\widetilde{x}_t=R(x_t,s_t).
\label{eq:scale-aware-resizing}
\end{equation}
The preprocessing operator \(R\) resizes each frame and aligns its dimensions with the embedding model's patch-size requirements.

Applying \(R\) frame-wise yields the mixed-resolution sequence
\begin{equation}
\widetilde{X}
=(\widetilde{x}_1,\ldots,\widetilde{x}_T).
\label{eq:allocated-video}
\end{equation}
Frames assigned larger scales preserve more spatial detail, whereas frames assigned smaller scales consume fewer visual resources.

\subsubsection{Embedding Backbone Integration}
The mixed-resolution sequence \(\widetilde{X}\) and task text \(q\) are passed to the existing embedding backbone \(E\), producing the retrieval embedding \(z\):
\begin{equation}
z=E(\widetilde{X},q).
\label{eq:video-embedding}
\end{equation}
Since allocation is completed before this stage, AllocEmbed leaves the backbone architecture, output dimensionality, and downstream similarity computation unchanged. This preserves compatibility with the original retrieval pipeline. By adaptively allocating visual detail, AllocEmbed enables the backbone to process more frames within the same visual budget, thereby further improving its retrieval performance.

\subsection{Retrieval-Driven Policy Optimization}
\label{sec:rdpo}

Existing datasets do not annotate the optimal input scale for each frame, and the utility of an allocation becomes observable only after the resulting frames have passed through preprocessing, embedding, and retrieval. We therefore introduce Retrieval-Driven Policy Optimization (RDPO), which learns the allocator directly from retrieval outcomes without frame-level scale supervision. Following recent task-feedback reinforcement learning methods~\citep{guo2025deepseek,yu2026dapo,jin2025tagging,zhang2026viper,jin2026finrpt}, RDPO builds on group-relative policy optimization (GRPO)~\citep{shao2024deepseekmath} and compares multiple allocations for the same input using an augmented retrieval candidate set, a rank-validated similarity gap, and an efficiency incentive that is gated by retrieval success and weighted by confidence. RDPO therefore trains the allocator using retrieval outcomes from the actual, potentially non-differentiable preprocessing and visual tokenization pipeline, without requiring a differentiable surrogate.

\subsubsection{Allocation Sampling}
For each training input \((X,q)\), RDPO samples a group of \(K\) frame-wise allocations from the allocator:
\begin{equation}
\mathbf{s}^{(1)},\ldots,\mathbf{s}^{(K)}
\sim\pi_{\theta}(\,\cdot\mid X,q).
\label{eq:allocation-samples}
\end{equation}
Each sampled allocation \(\mathbf{s}^{(k)}\) is applied to \(X\) through scale-aware preprocessing, and the frozen embedding backbone encodes the resulting mixed-resolution sequence together with \(q\) to produce \(z_k\). Within the group, the video, task text, positive target, and retrieval candidates remain fixed, while only the sampled allocation varies. Their retrieval outcomes therefore provide a controlled comparison of how frame-wise resource allocation affects retrieval quality for that input.

\subsubsection{Global Negative Sample Augmentation}
Because in-batch negatives alone may be insufficiently challenging, we supplement them with text negatives sampled from the training corpus to form the candidate set
\begin{equation}
\mathcal{U}
=\{u^+\}\cup\mathcal{N}^{\mathrm{batch}}
\cup\mathcal{N}^{\mathrm{global}},
\label{eq:candidate-set}
\end{equation}
where \(u^+\) is the positive target embedding, and \(\mathcal{N}^{\mathrm{batch}}\) and \(\mathcal{N}^{\mathrm{global}}\) are the in-batch and global negative embedding sets, respectively. All \(K\) allocations are evaluated against the same \(\mathcal{U}\), ensuring a controlled comparison. Since the added negatives are text-only, their encoding cost is small relative to video encoding.

\subsubsection{Rank-Validated Similarity Gap}
For the \(k\)-th allocation, RDPO evaluates retrieval success and the positive target's separation from the negatives. The retrieval success indicator is
\begin{equation}
a_k=\mathbf{1}[\mathrm{rank}_k(u^+)=1].
\label{eq:retrieval-success}
\end{equation}
For the negative set \(\mathcal{N}=\mathcal{U}\setminus\{u^+\}\), the mean negative similarity is
\begin{equation}
\mu_k^-=
\frac{1}{|\mathcal{N}|}
\sum_{u^-\in\mathcal{N}}
\mathrm{sim}(z_k,u^-),
\label{eq:mean-negative-similarity}
\end{equation}
and the positive-negative similarity gap is
\begin{equation}
g_k=\mathrm{sim}(z_k,u^+)-\mu_k^-.
\label{eq:similarity-gap}
\end{equation}
RDPO gates the similarity gap by retrieval success to obtain the rank-validated similarity gap
\begin{equation}
r_k=a_kg_k.
\label{eq:rank-validated-gap}
\end{equation}
This assigns a score of zero to unsuccessful retrievals and grades successful ones by their separation from the negatives.

We obtain the group-relative retrieval advantage by normalizing \(r_k\) across the \(K\) allocations of the same input:
\begin{equation}
\scalebox{0.82}{$\displaystyle
A_{\mathrm{ret},k}
=\frac{r_k-\bar{r}}
{\mathrm{std}(r_1,\ldots,r_K)+\epsilon},
\quad
\bar{r}=\frac{1}{K}\sum_{j=1}^{K}r_j.
$}
\label{eq:group-relative-retrieval-advantage}
\end{equation}
This normalization compares allocation quality within each input, reducing sensitivity to differences in example difficulty.

\subsubsection{Confidence-Guided Efficiency Incentive}
RDPO additionally encourages efficient allocations, but only when they already retrieve the correct target with sufficient confidence. For the \(k\)-th allocation, define the mean scale and its normalized cost proxy as
\begin{equation}
\scalebox{0.82}{$\displaystyle
\bar{s}_k=\frac{1}{T}\sum_{t=1}^{T}s_t^{(k)},\quad
c_k=\mathrm{clip}\left(
\frac{\bar{s}_k-s_{\min}}{s_{\max}-s_{\min}},0,1
\right).
$}
\label{eq:mean-scale-cost}
\end{equation}
We derive a soft confidence score from the similarity gap:
\begin{equation}
h_k=\sigma\left(\frac{g_k-\gamma}{\tau}\right),
\label{eq:confidence-score}
\end{equation}
where \(\gamma\) is a confidence threshold and \(\tau\) controls the softness of the transition. The efficiency incentive is
\begin{equation}
I_{\mathrm{eff},k}
=\lambda_{\mathrm{eff}}a_kh_k(1-c_k).
\label{eq:efficiency-incentive}
\end{equation}
This term is an auxiliary incentive conditioned on retrieval success, rather than an independent retrieval reward. An unsuccessful allocation cannot benefit merely by using a small scale. A successful but small-gap allocation receives only a weak efficiency incentive, whereas a successful and confident allocation is encouraged to reduce its average scale. The final advantage is
\begin{equation}
A_k=A_{\mathrm{ret},k}+I_{\mathrm{eff},k}.
\label{eq:final-advantage}
\end{equation}

\subsubsection{Policy Optimization}
For the \(k\)-th allocation, RDPO assigns \(A_k\) to every valid frame action. Let \(\pi_{\theta_{\mathrm{old}}}\) denote the allocator policy used to sample the group of allocations. The likelihood ratio for the normalized Beta action \(a_t^{(k)}\) is
\begin{equation}
\rho_{k,t}=
\frac{\pi_{\theta}(a_t^{(k)}\mid X,q)}
{\pi_{\theta_{\mathrm{old}}}(a_t^{(k)}\mid X,q)}.
\label{eq:rdpo-policy-ratio}
\end{equation}
RDPO updates the allocator by maximizing the following clipped PPO objective~\citep{schulman2017proximal}:
\begin{equation}
\scalebox{0.9}{$\displaystyle
\mathcal{J}(\theta)
=\mathbb{E}_{k,t}\!\left[
\min\!\left(\rho_{k,t}A_k,
\mathrm{clip}_{l,h}(\rho_{k,t})A_k\right)
\right].
$}
\label{eq:rdpo-policy-objective}
\end{equation}
Here, \(\mathrm{clip}_{l,h}(\rho)=\mathrm{clip}(\rho,1-\epsilon_l,1+\epsilon_h)\), and the expectation is over the sampled allocations and their valid frame actions. The embedding backbone remains frozen throughout optimization.
\begin{table*}[t]
\centering
{\footnotesize
\setlength{\tabcolsep}{1.4pt}
\renewcommand{\arraystretch}{1.12}
\resizebox{\textwidth}{!}{%
\begin{tabular}{@{}lcccccccccccc@{}}
  \toprule
  Method &
  \multicolumn{5}{c}{V-RET} &
  \multicolumn{5}{c}{V-QA} &
  Avg. &
  Cost \\
  \cmidrule(lr){2-6}\cmidrule(lr){7-11}
  & VTT & MSVD & DiDeMo & YC2 & VATEX
  & V-MME & NExT & EgoSch. & MVB & ANetQA
  & & \\
  \midrule
  \multicolumn{13}{@{}l}{\textbf{VLM2Vec-V2-2B}} \\
  Base
  & $28.30$ & $48.06$ & $30.38$ & $10.63$ & $26.46$
  & $30.70$ & $20.92$ & $34.00$ & $33.70$ & $52.30$
  & $31.55$ & $1.000{\times}$ \\
  Full Resolution$^\dagger$
  & $31.90$ & $50.15$ & $33.07$ & $11.20$ & $30.77$
  & $31.07$ & $21.28$ & $36.80$ & $34.15$ & $49.80$
  & $33.02\,(+1.47)$ & $2.952{\times}$ \\
  Uniform Allocation
  & $25.80$ & $\mathbf{49.10}$ & $31.87$ & $11.61$ & $28.20$
  & $32.26$ & $21.54$ & $34.80$ & $34.25$ & $51.90$
  & $32.13\,(+0.58)$ & $1.025{\times}$ \\
  Content-Aware Allocation
  & $26.40$ & $45.67$ & $31.87$ & $11.48$ & $27.78$
  & $32.52$ & $21.19$ & $35.80$ & $33.50$ & $\mathbf{53.70}$
  & $31.99\,(+0.44)$ & $1.003{\times}$ \\
  Content-Aware Selection
  & $26.80$ & $45.22$ & $30.68$ & $9.81$ & $25.10$
  & $31.07$ & $20.95$ & $32.20$ & $33.62$ & $51.20$
  & $30.67\,(-0.88)$ & $1.000{\times}$ \\
  \textbf{AllocEmbed}
  & $\mathbf{29.00}$ & $48.66$ & $\mathbf{32.77}$ & $\mathbf{12.05}$ & $\mathbf{28.94}$
  & $\mathbf{32.56}$ & $\mathbf{21.82}$ & $\mathbf{38.60}$ & $\mathbf{34.48}$ & $53.20$
  & $\mathbf{33.21\,(+1.66)}$ & $0.943{\times}$ \\
  \midrule
  \multicolumn{13}{@{}l}{\textbf{Qwen3-VL-Embedding-2B}} \\
  Base
  & $47.00$ & $68.21$ & $43.33$ & $23.28$ & $40.31$
  & $50.48$ & $64.85$ & $57.80$ & $57.63$ & $73.60$
  & $52.65$ & $1.000{\times}$ \\
  Full Resolution$^\dagger$
  & $51.10$ & $71.49$ & $48.21$ & $28.22$ & $45.40$
  & $53.56$ & $67.90$ & $62.40$ & $59.17$ & $75.40$
  & $56.28\,(+3.63)$ & $2.923{\times}$ \\
  Uniform Allocation
  & $47.20$ & $67.91$ & $46.91$ & $25.86$ & $41.27$
  & $53.93$ & $\mathbf{65.30}$ & $57.60$ & $\mathbf{57.95}$ & $73.70$
  & $53.76\,(+1.11)$ & $1.002{\times}$ \\
  Content-Aware Allocation
  & $46.00$ & $68.66$ & $45.92$ & $24.91$ & $41.02$
  & $52.93$ & $65.26$ & $54.80$ & $56.60$ & $\mathbf{74.00}$
  & $53.01\,(+0.36)$ & $1.038{\times}$ \\
  Content-Aware Selection
  & $46.30$ & $65.67$ & $45.82$ & $22.84$ & $39.01$
  & $50.93$ & $65.14$ & $\mathbf{58.40}$ & $56.95$ & $72.40$
  & $52.35\,(-0.30)$ & $1.000{\times}$ \\
  \textbf{AllocEmbed}
  & $\mathbf{50.20}$ & $\mathbf{69.70}$ & $\mathbf{47.61}$ & $\mathbf{27.27}$ & $\mathbf{43.52}$
  & $\mathbf{54.85}$ & $63.70$ & $55.80$ & $56.97$ & $73.70$
  & $\mathbf{54.33\,(+1.68)}$ & $0.997{\times}$ \\
  \bottomrule
\end{tabular}%
}
}
\caption{MMEB-V2 Hit@1 (\%) on ten video tasks with two 2B backbones. Base embeds 8 uniformly sampled frames; other methods use 24, except Content-Aware Selection, which embeds 8 selected frames. Avg. is the task mean, with its change from Base in parentheses, and Cost is normalized to Base ($1.000{\times}$). Bold marks the best budget-matched result per column. $^\dagger$ Full Resolution uses 24 frames at Base resolution and is not budget matched. The task columns are MSR-VTT, MSVD, DiDeMo, YouCook2, VATEX, Video-MME, NExTQA, EgoSchema, MVBench, and ActivityNetQA.}
\label{tab:main-mmeb10-2b}
\vspace{-1.5pt}
\end{table*}

\section{Experiments}
\label{sec:experiments}

\subsection{Experimental Setup}
\label{sec:experimental-setup}

\paragraph{Implementation Details.}
We train AllocEmbed on the video training set used by VLM2Vec-V2~\citep{meng2025vlm2vec} while keeping both the SmolVLM2-256M feature extractor and the VLM2Vec-V2-2B embedding backbone frozen. Only the frame-wise allocation module is optimized for 125 steps on eight NVIDIA H100 80GB GPUs with a learning rate of \(2\times10^{-5}\). We use a batch size of 1,024 video--text examples, sampling \(K=16\) allocations per example with continuous scales in \([0.2,1.8]\). We set \((\lambda_{\mathrm{eff}},\gamma,\tau)=(0.4,0.45,0.10)\) and add 3,072 global text negatives. Further training and RDPO details are provided in the appendix.

\paragraph{Benchmarks and Baselines.}
We evaluate AllocEmbed on ten MMEB-V2 video tasks~\citep{meng2025vlm2vec} and LongRet, a newly constructed benchmark for long-video retrieval. We report budget-matched Hit@1 unless noted otherwise. Base encodes eight uniformly sampled frames at their original resolution, while the other methods begin with the same 24 uniformly sampled frames. Uniform Allocation, Content-Aware Allocation, and AllocEmbed retain all 24 but allocate resolution differently, whereas Content-Aware Selection embeds eight selected frames. All methods match the Base budget except Full Resolution, which processes all 24 at the Base resolution as a high-cost reference. The appendix provides benchmark and baseline details.

\subsection{Main Results}
\label{sec:main-results}

\label{sec:overall-results}

Table~\ref{tab:main-mmeb10-2b} reports task-level and average MMEB-V2 results together with realized costs. AllocEmbed achieves the best budget-matched average with both 2B models while matching or reducing the Base cost. It thus better balances temporal coverage and spatial fidelity. This confirms the value of learned frame-wise resolution allocation.

\begin{figure}[t]
\centering
\includegraphics[width=\columnwidth]{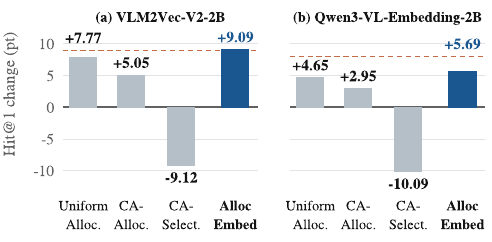}
\caption{Average LongRet Hit@1 gains over the 8-frame Base. CA-Alloc. and CA-Select. denote Content-Aware Allocation and Selection. Bars are budget matched; dashed lines show Full Resolution at roughly \(3\times\) Base cost.}
\label{fig:main-longret-2b}
\vspace{-1.5pt}
\end{figure}

Figure~\ref{fig:main-longret-2b} summarizes performance on the four LongRet subsets, with complete results provided in the appendix. AllocEmbed's gains over Base are more pronounced on LongRet, ranging from 5.69 to 9.09 Hit@1 points, as sparse frame sampling is more likely to miss retrieval-relevant evidence in long videos. It is the strongest budget-matched method on all four subsets and recovers most of Full Resolution's improvement at roughly one-third of the cost. This highlights the importance of preserving broad temporal coverage while varying spatial fidelity in long-video retrieval.

\subsection{Ablation Studies}
\label{sec:ablation-studies}

\label{sec:rdpo-ablation}

We ablate RDPO across the ten MMEB-V2 video tasks using VLM2Vec-V2-2B, retraining each variant on the same training data. During evaluation, all variants use 24 input frames, with the remaining allocation settings unchanged from training. Table~\ref{tab:rdpo-ablation-vlm2vec} reports average Hit@1 and realized visual cost. Full RDPO achieves the best quality--cost balance. Retrieval success alone provides too sparse a learning signal, while the ungated similarity gap preserves most of the accuracy at greater visual cost. Combining retrieval success with the similarity gap provides both a reliable success criterion and graded ranking feedback. Removing global negatives causes the largest accuracy loss of 1.30 points, highlighting the importance of challenging negatives beyond in-batch comparisons. Confidence guidance, success gating, and the efficiency incentive primarily regulate cost: removing any of them offers little benefit while increasing resource use.

\begin{table}[t]
\centering
\footnotesize
\setlength{\tabcolsep}{2pt}
\renewcommand{\arraystretch}{1.08}
\begin{tabular}{@{}lrrrr@{}}
  \toprule
  Variant & Hit@1 & $\Delta$ (pt) & Cost & $\Delta$ (\%) \\
  \midrule
  \textbf{Full RDPO} & $33.21$ & --- & $2{,}008.44$ & --- \\
  Retrieval Success Only & $32.05$ & $-1.16$ & $1{,}576.40$ & $-21.51$ \\
  Similarity Gap Only & $32.92$ & $-0.29$ & $2{,}138.27$ & $+6.46$ \\
  w/o Global Negatives & $31.91$ & $-1.30$ & $1{,}242.87$ & $-38.12$ \\
  w/o Confidence Guidance & $32.90$ & $-0.31$ & $2{,}155.19$ & $+7.31$ \\
  w/o Efficiency Incentive & $32.85$ & $-0.36$ & $2{,}180.98$ & $+8.59$ \\
  w/o Success Gating & $33.05$ & $-0.16$ & $2{,}452.24$ & $+22.10$ \\
  \bottomrule
\end{tabular}
\caption{RDPO training-signal ablation on ten MMEB-V2 tasks with VLM2Vec-V2-2B. Values are mean Hit@1 (\%), and both \(\Delta\) columns are relative to Full RDPO.}
\label{tab:rdpo-ablation-vlm2vec}
\vspace{-1.5pt}
\end{table}

\subsection{Analysis and Discussion}
\label{sec:analysis-discussion}

\subsubsection{Generalization to Diverse Embedding Models}
\label{sec:backbone-scaling}

To assess cross-model transfer, we evaluate AllocEmbed with four embedding models spanning 2B to 8B parameters: VLM2Vec-V2-2B, VLM2Vec-Qwen2VL-7B, Qwen3-VL-Embedding-2B, and Qwen3-VL-Embedding-8B. Trained only with VLM2Vec-V2-2B, the allocator transfers directly to the other three models without further training. Table~\ref{tab:backbone-scaling-summary} reports the Hit@1 gain and visual cost relative to each model's own Base, using the same MMEB-V2 and LongRet averages as above. Across all four models, AllocEmbed improves both averages while matching or reducing the visual cost of the corresponding Base. The gains are consistent on MMEB-V2 and substantially larger on LongRet, showing that the learned allocation policy transfers across embedding architectures and parameter scales and remains effective when retrieval depends on sparse evidence in long videos.

\begin{table}[t]
\centering
\footnotesize
\setlength{\tabcolsep}{2.2pt}
\renewcommand{\arraystretch}{1.08}
\begin{tabular*}{\columnwidth}{@{\extracolsep{\fill}}lrrrr@{}}
  \toprule
  Model & \multicolumn{2}{c}{V-RET \& V-QA} & \multicolumn{2}{c}{LongRet} \\
  \cmidrule(lr){2-3}\cmidrule(lr){4-5}
  & Gain & Cost & Gain & Cost \\
  \midrule
  VLM2Vec-V2-2B & $+1.66$ & $0.943{\times}$ & $+9.09$ & $0.979{\times}$ \\
  VLM2Vec-Qwen2VL-7B & $+1.34$ & $0.943{\times}$ & $+7.07$ & $0.979{\times}$ \\
  Qwen3-VL-Emb.-2B & $+1.68$ & $0.997{\times}$ & $+5.69$ & $0.977{\times}$ \\
  Qwen3-VL-Emb.-8B & $+1.69$ & $0.997{\times}$ & $+7.63$ & $0.977{\times}$ \\
  \bottomrule
\end{tabular*}
\caption{Generalization across embedding models. Gain is the Hit@1 change from the 8-frame Base, and Cost is visual cost normalized to Base.}
\label{tab:backbone-scaling-summary}
\vspace{-1.5pt}
\end{table}

\subsubsection{Performance across Visual Input Budgets}
\label{sec:budget-analysis}

\begin{figure}[t]
\centering
\includegraphics[width=0.96\columnwidth]{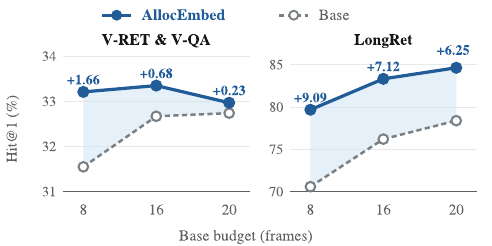}
\caption{Performance across visual-input budgets with VLM2Vec-V2-2B. Base uses 8, 16, or 20 native-resolution frames; AllocEmbed uses 24, 48, or 60 candidates under matched Base budgets. Curves show mean Hit@1 on ten MMEB-V2 tasks (left) and four LongRet subsets (right); labels give absolute gains over Base.}
\label{fig:visual-budget-scaling}
\vspace{-1.5pt}
\end{figure}

To assess robustness across visual input budgets, we evaluate AllocEmbed with VLM2Vec-V2-2B at three budget levels. Base processes 8, 16, or 20 native-resolution frames, while AllocEmbed uses candidate pools of 24, 48, or 60 frames under the corresponding Base budget. Figure~\ref{fig:visual-budget-scaling} shows that AllocEmbed outperforms Base at every level on both MMEB-V2 and LongRet. Gains are especially pronounced on LongRet, reaching 6.25 Hit@1 points at the largest 20-frame budget. These results show that learned allocation remains effective as both the Base frame count and candidate pool scale up. Further analysis of inference efficiency is provided in the appendix.

\subsubsection{Case Study}
\label{sec:case-study}

To inspect AllocEmbed's allocations, we analyze two examples with complete 24-frame visualizations in the appendix. For the Oscars location question, clear frames with the red carpet, statuettes, and location-revealing text receive larger scales, while a black transition receives the minimum. For character counting, sharp game footage and text-bearing frames receive larger scales than blank or blurred transitions. These examples show that AllocEmbed prioritizes informative, high-quality frames.

\section{Conclusion}

We introduced AllocEmbed, an allocate-then-embed framework that balances temporal coverage and spatial fidelity by distributing a constrained visual-input budget across a denser frame sequence. AllocEmbed predicts input resolutions from video content and task text before embedding, allowing existing VLM-based embedding models to process mixed-resolution videos without architectural changes. RDPO learns this allocation directly from retrieval outcomes through a rank-validated similarity gap and a confidence-guided efficiency incentive. Experiments on MMEB-V2 and our newly constructed LongRet benchmark show improved aggregate retrieval performance under matched visual-input budgets, together with transfer across backbones from 2B to 8B parameters without retraining. These results highlight input-side visual allocation as a practical approach to improving video retrieval with existing embedding backbones.

\bibliography{custom}

\clearpage
\appendix
\section{Additional Experimental Setup}
\label{app:additional-experimental-setup}

\subsection{Implementation Details}
\label{app:implementation-details}

\begin{table*}[!t]
\centering
\footnotesize
\renewcommand{\arraystretch}{1.08}
\begin{tabular}{@{}p{0.12\textwidth}@{\hspace{8pt}}p{0.20\textwidth}@{\hspace{8pt}}p{0.22\textwidth}@{\hspace{8pt}}p{0.40\textwidth}@{}}
\toprule
Category & Setting & Value & Notes \\
\midrule
Infrastructure & Hardware & 8 NVIDIA H100 80GB GPUs & Distributed training on a single node. \\
Infrastructure & Parallelism and precision & FSDP2, BF16 & FSDP2 parameter sharding with BF16 computation. \\
\midrule
Model & Embedding backbone & VLM2Vec-V2-2B (\(\sim\)2.21B) & Frozen model used to compute embeddings and retrieval rewards. \\
Model & Allocator implementation & \shortstack[l]{SmolVLM2-256M-based\\allocator} & Frozen SmolVLM2-256M feature extractor followed by a trainable frame-wise scale predictor. \\
\midrule
Optimization & Allocator learning rate & \(2\times10^{-5}\) & Applied only to the trainable scale predictor. \\
Optimization & Training duration & 125 steps & Total number of training steps. \\
\midrule
Data & Video splits & \shortstack[l]{Complete training split;\\complete validation split} & Train on the training split; use the validation split for checkpoint selection. \\
\midrule
Batching & Examples per step & 1,024 & Video--text examples sampled before allocation expansion. \\
Batching & Allocations per example & \(K=16\) & Scale allocations sampled for each example. \\
\midrule
Allocation & Scale range & \([0.2,1.8]\) & Per-frame scale applied to both height and width. \\
\midrule
RDPO & Additional global negatives & 3,072 & Text targets sampled beyond the current batch. \\
RDPO & Efficiency weight & \(\lambda_{\mathrm{eff}}=0.4\) & Strength of the success-gated efficiency incentive. \\
RDPO & Confidence gate & \(\gamma=0.45,\ \tau=0.10\) & Center and temperature for similarity-gap confidence. \\
RDPO & Clipping widths & \(\epsilon_l=0.20,\ \epsilon_h=0.28\) & Asymmetric policy-ratio clipping widths. \\
\bottomrule
\end{tabular}
\caption{Training configuration used for AllocEmbed. Only the frame-wise scale predictor is trained, while the SmolVLM2 feature extractor and embedding backbone remain frozen.}
\label{tab:training-hyperparameters}
\end{table*}

Table~\ref{tab:training-hyperparameters} summarizes the training configuration. We train on the complete VLM2Vec-V2 video training split and use the complete validation split for checkpoint selection; neither split is manually truncated. The VLM2Vec-V2-2B embedding backbone contains approximately 2.21B parameters and remains frozen, providing the embeddings used to compute retrieval rewards. The allocator encodes candidate-frame previews and task text with a frozen SmolVLM2-256M feature extractor. Only the attached frame-wise scale predictor is optimized.

Training runs on a single node with eight NVIDIA H100 80GB GPUs using FSDP2 and BF16 computation. We optimize the scale predictor for 125 steps with a learning rate of \(2\times10^{-5}\). Each step samples 1,024 video--text examples and draws \(K=16\) allocations per example, yielding 16,384 allocation-conditioned queries. The prompt mini-batch size is 256. After allocation expansion and distributed sharding, the allocator mini-batch contains 512 samples per GPU. Scale prediction and RDPO backpropagation use a micro-batch size of 64 per GPU, while the frozen embedding backbone encodes queries and targets in batches of 128 per GPU.

For each candidate frame, the predictor outputs a continuous scale in \([0.2,1.8]\). Because the scale is applied to both spatial dimensions, visual input cost grows approximately quadratically with scale.

Following the RDPO formulation in the main text, we set \(\lambda_{\mathrm{eff}}=0.4\), \(\gamma=0.45\), and \(\tau=0.10\). The success gate prevents an incorrect allocation from receiving an efficiency incentive solely by reducing its scale. We augment the 1,024 in-batch targets with 3,072 text targets sampled from the VLM2Vec-V2 training corpus. The resulting 4,096-target pool is shared across all 16,384 allocation-conditioned queries. Additional negatives exclude the current positive, targets with the same sample identity, and exact text duplicates.

\subsection{Benchmark and Baseline Details}
\label{app:benchmark-baseline-details}

Our MMEB-V2 evaluation comprises five video-retrieval (V-RET) tasks and five video-question-answering (V-QA) tasks. The V-RET tasks are MSR-VTT~\citep{xu2016msr}, MSVD~\citep{chen2011collecting}, DiDeMo~\citep{anne2017localizing}, YouCook2~\citep{zhou2018towards}, and VATEX~\citep{wang2019vatex}. The V-QA tasks are Video-MME~\citep{fu2025video}, NExTQA~\citep{xiao2021next}, EgoSchema~\citep{mangalam2023egoschema}, MVBench~\citep{li2024mvbench}, and ActivityNet-QA~\citep{yu2019activitynet}. We evaluate four embedding models: VLM2Vec-V2-2B, Qwen3-VL-Embedding-2B, VLM2Vec-Qwen2VL-7B, and Qwen3-VL-Embedding-8B.

\paragraph{LongRet benchmark construction.}
LongRet evaluates text-to-video retrieval at two query granularities while using complete source videos as retrieval targets. Its candidate corpora contain 467 LoVR videos~\citep{liang2026lovr} and 2,375 Vript videos~\citep{yang2024vript}. Localized scene annotations contribute query text only: no scene clip is treated as a retrieval target, and every query must retrieve the complete source video from which its description originates. We therefore compute all duration statistics over unique complete-video targets rather than over localized scene annotations.

The Clip subsets use localized descriptions. We randomly sample up to three eligible segment captions per video and treat each caption as a separate query, producing 1,401 LoVR-Clip queries and 7,112 Vript-Clip queries. The Video subsets use complete-video descriptions. LoVR provides one video-level caption per source video, yielding 467 LoVR-Video queries. For Vript, we concatenate the clip captions associated with each source video into one complete-video query, yielding 2,375 Vript-Video queries. Overall, LongRet contains 11,355 queries over 2,842 source videos. Table~\ref{tab:longret-composition} reports the evaluation composition, full-video duration statistics, and duration-bin counts.

\begin{table*}[!t]
\centering
{\footnotesize
\setlength{\tabcolsep}{3.8pt}
\renewcommand{\arraystretch}{1.08}
\begin{tabular*}{\textwidth}{@{\extracolsep{\fill}}lrrrl@{}}
\multicolumn{5}{@{}l}{\textbf{(a) Evaluation composition}} \\
\toprule
Subset & Queries & \shortstack{Unique full-video\\targets} & \shortstack{Source\\videos} & Query--target mapping \\
\midrule
LoVR-Clip  & 1,401 & 467   & 467   & Local description $\rightarrow$ complete video \\
LoVR-Video & 467   & 467   & 467   & Complete description $\rightarrow$ complete video \\
Vript-Clip & 7,112 & 2,375 & 2,375 & Local description $\rightarrow$ complete video \\
Vript-Video& 2,375 & 2,375 & 2,375 & Complete description $\rightarrow$ complete video \\
\bottomrule
\end{tabular*}

\vspace{5pt}
{\footnotesize
\setlength{\tabcolsep}{2.5pt}
\begin{tabular*}{\textwidth}{@{\extracolsep{\fill}}lrrrrrrrrr@{}}
\multicolumn{10}{@{}l}{\textbf{(b) Unique complete-video duration statistics}} \\
\toprule
Corpus & Videos & \shortstack{Total\\(h)} & \multicolumn{7}{c}{Duration (min)} \\
\cmidrule(lr){4-10}
& & & Mean & Min & P25 & Median & P75 & P95 & Max \\
\midrule
LoVR  & 467   & 202.41 & 26.01 & 16.64 & 18.49 & 21.56 & 29.00 & 50.59 & 59.57 \\
Vript & 2,375 & 563.53 & 14.24 & 10.00 & 11.28 & 13.39 & 16.36 & 20.93 & 29.82 \\
\bottomrule
\end{tabular*}
}

\vspace{5pt}
\begin{tabular*}{\textwidth}{@{\extracolsep{\fill}}lrrrrr@{}}
\multicolumn{6}{@{}l}{\textbf{(c) Unique complete-video duration distribution}} \\
\toprule
Corpus & \(<10\) min & 10--15 min & 15--20 min & 20--30 min & \(\geq 30\) min \\
\midrule
LoVR  & 0 & 0 & 189 (40.5\%) & 170 (36.4\%) & 108 (23.1\%) \\
Vript & 0 & 1,560 (65.7\%) & 661 (27.8\%) & 154 (6.5\%) & 0 \\
\bottomrule
\end{tabular*}
}
\caption{LongRet composition and complete-video duration statistics. Clip and Video share targets and duration distributions within each corpus but differ in query granularity and count. Bins are left-closed and right-open, except the final bin.}
\label{tab:longret-composition}
\end{table*}

\paragraph{Baseline and evaluation protocols.}
Under the default protocol, Base embeds 8 uniformly sampled frames at the original resolution, while the other methods begin with 24 candidate frames. Full Resolution processes all 24 frames at the Base resolution and incurs a higher visual cost. Uniform Allocation lowers their resolution uniformly to match the Base budget. Content-Aware Allocation assigns resolutions using a frame-importance heuristic, while Content-Aware Selection embeds the 8 highest-scoring frames. AllocEmbed predicts a continuous, task-conditioned scale for each candidate frame.

Across methods, we hold the input video, task text, retrieval targets, and metric fixed, and report Hit@1 with realized visual input cost. All methods except Full Resolution are evaluated under matched budgets. At inference, AllocEmbed predicts one allocation and invokes the frozen backbone once after scale-aware preprocessing. Group sampling with \(K=16\) is used only for training.

\FloatBarrier
\section{Additional Analysis and Discussion}
\label{app:additional-analysis-discussion}

\subsection{Case Study Visualizations}
\label{app:case-study-visualizations}

\begin{figure*}[!t]
  \centering
  \includegraphics[width=\textwidth]{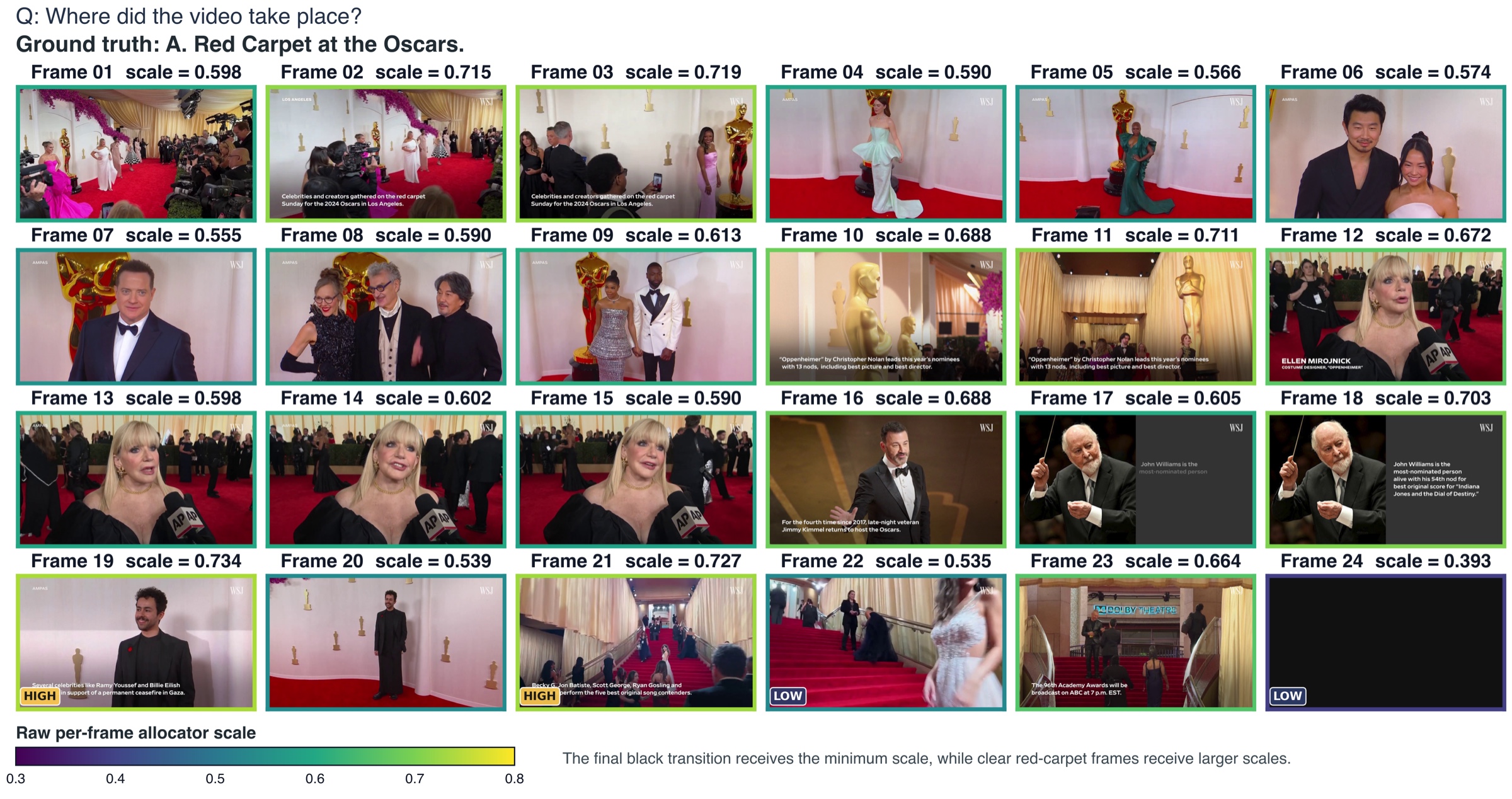}
  \caption{Frame-level scales for a location question. AllocEmbed assigns the minimum scale to the final black transition and larger scales to clear views of the red carpet and Oscar iconography.}
  \label{fig:case-red-carpet}
\end{figure*}

\begin{figure*}[!t]
  \centering
  \includegraphics[width=\textwidth]{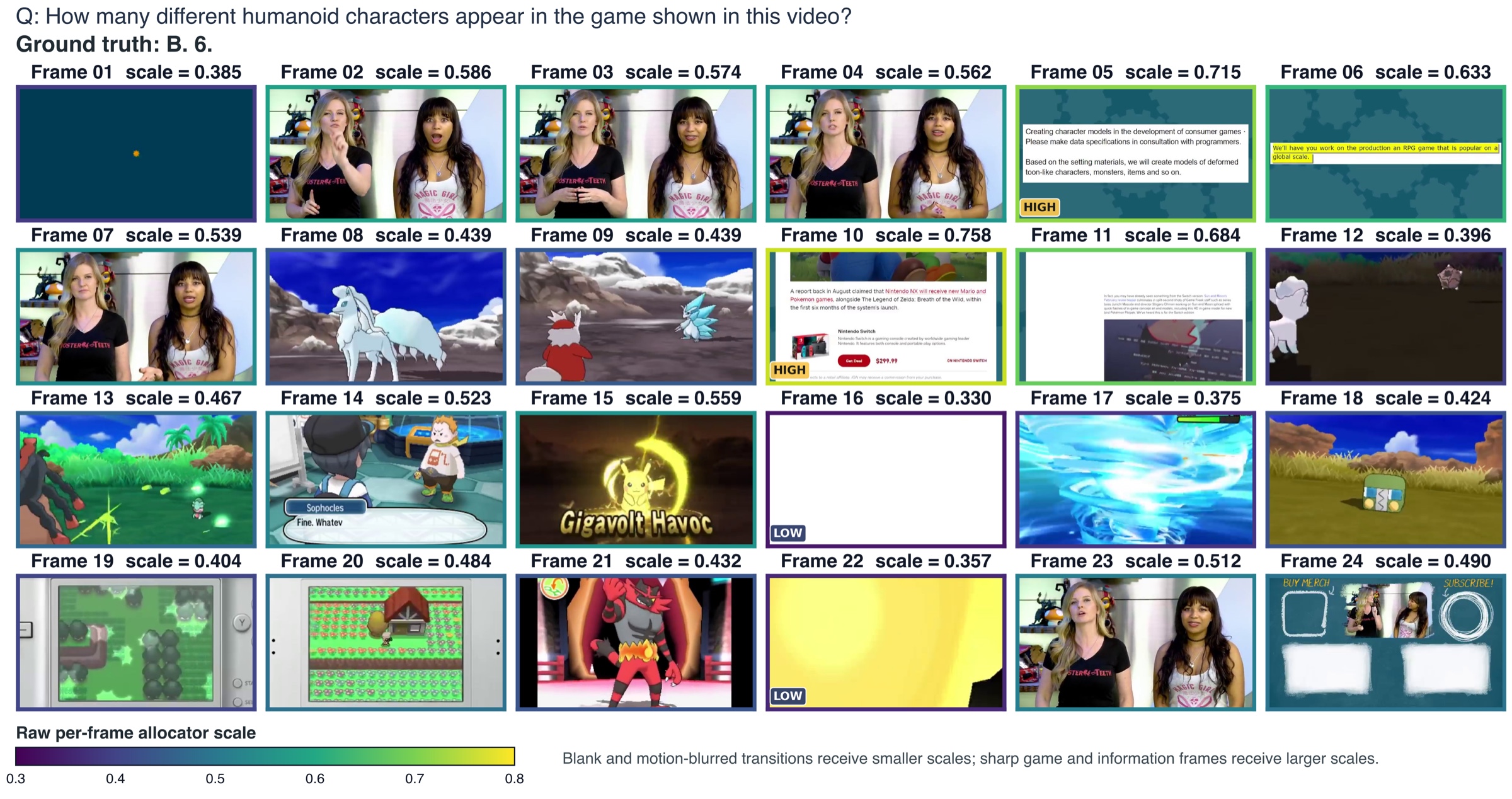}
  \caption{Frame-level scales for a character-counting question. AllocEmbed assigns smaller scales to blank and motion-blurred transitions and larger scales to sharp game footage and text-rich frames.}
  \label{fig:case-game}
\end{figure*}

Figures~\ref{fig:case-red-carpet} and~\ref{fig:case-game} present the complete 24-frame visualizations for the examples discussed in the Case Study subsection. Each frame is labeled with its raw predicted scale, and the border color encodes the same value.

\subsection{Inference Overhead}
\label{app:inference-overhead}

We measure inference overhead on the LoVR-Video candidate videos using VLM2Vec-V2.0 (Qwen2-VL). All experiments run in BF16 with batch size one on a single NVIDIA H100 80GB GPU. In paired runs, AllocEmbed and the full-input baseline process the same videos, text, and 24 sampled frames in the same order under a 64K video-token cap. AllocEmbed predicts scales in two-frame chunks within \([0.2,1.32]\). The full-input baseline bypasses allocation and sends all 24 frames directly to the backbone.

We discard five warm-up samples and average latency over the remaining 462 videos, synchronizing CUDA around each timed region. Allocator latency includes text preparation, host-to-device transfer, the allocator encoders and prediction head, and scale generation. Backbone latency includes host-to-device transfer and one VLM2Vec forward pass. Gradient computation and parameter updates are disabled for all runs.

As shown in Table~\ref{tab:inference-overhead}, the allocator adds \(149.59\) ms per sample but reduces backbone latency from \(754.77\) to \(239.22\) ms. Total model-path latency decreases by \(48.49\%\), corresponding to a \(1.941\times\) speedup over full-input inference. These measurements exclude allocator preprocessing, scale-dependent resizing, video transforms, and backbone preprocessing. The excluded stages can run on the CPU and are amenable to further systems optimization through parallelism and pipelining. A full optimization of these stages is outside the scope of this work.

\begin{table}[t]
\centering
\footnotesize
\setlength{\tabcolsep}{2pt}
\renewcommand{\arraystretch}{1.08}
\begin{tabular*}{\columnwidth}{@{\extracolsep{\fill}}lrrrr@{}}
  \toprule
  & & \multicolumn{3}{c}{Latency (ms/sample)} \\
  \cmidrule(lr){3-5}
  Method & Params & Allocator & Backbone & Total \\
  \midrule
  Full input & $2.218$B & -- & $754.77$ & $754.77$ \\
  \textbf{AllocEmbed} & $2.523$B & $149.59$ & $239.22$ & $\mathbf{388.81}$ \\
  \bottomrule
\end{tabular*}
\caption{Model-path latency on LoVR-Video. Results are averaged over 462 samples after five warm-up samples. Params denotes the total number of loaded parameters.}
\label{tab:inference-overhead}
\end{table}

\subsection{Complete LongRet Results}
\label{app:complete-longret-results}

Table~\ref{tab:main-longret-2b} reports the complete subset-level LongRet results summarized in Figure~\ref{fig:main-longret-2b}.

\begin{table*}[!t]
\centering
{\footnotesize
\setlength{\tabcolsep}{2pt}
\renewcommand{\arraystretch}{1.12}
\begin{tabular*}{\textwidth}{@{\extracolsep{\fill}}lcccccc@{}}
  \toprule
  Method &
  \multicolumn{2}{c}{LoVR} &
  \multicolumn{2}{c}{Vript} &
  Avg. &
  Cost \\
  \cmidrule(lr){2-3}\cmidrule(lr){4-5}
  & Clip & Video & Clip & Video & & \\
  \midrule
  \multicolumn{7}{@{}l}{\textbf{VLM2Vec-V2-2B}} \\
  Base
  & $61.10$ & $74.09$ & $69.92$ & $77.26$
  & $70.59$ & $1.000{\times}$ \\
  Full Resolution$^\dagger$
  & $71.23$ & $83.73$ & $78.63$ & $84.59$
  & $79.54\,(+8.95)$ & $2.997{\times}$ \\
  Uniform Allocation
  & $70.45$ & $82.66$ & $75.58$ & $84.76$
  & $78.36\,(+7.77)$ & $1.016{\times}$ \\
  Content-Aware Allocation
  & $65.67$ & $80.09$ & $72.60$ & $84.21$
  & $75.64\,(+5.05)$ & $1.001{\times}$ \\
  Content-Aware Selection
  & $48.39$ & $65.10$ & $58.96$ & $73.43$
  & $61.47\,(-9.12)$ & $1.000{\times}$ \\
  \textbf{AllocEmbed}
  & $\mathbf{71.45}$ & $\mathbf{83.73}$ & $\mathbf{77.24}$ & $\mathbf{86.32}$
  & $\mathbf{79.68\,(+9.09)}$ & $0.979{\times}$ \\
  \midrule
  \multicolumn{7}{@{}l}{\textbf{Qwen3-VL-Embedding-2B}} \\
  Base
  & $80.94$ & $83.51$ & $81.83$ & $88.51$
  & $83.70$ & $1.000{\times}$ \\
  Full Resolution$^\dagger$
  & $91.01$ & $90.58$ & $91.84$ & $93.43$
  & $91.72\,(+8.02)$ & $2.984{\times}$ \\
  Uniform Allocation
  & $82.87$ & $89.51$ & $88.05$ & $92.97$
  & $88.35\,(+4.65)$ & $1.036{\times}$ \\
  Content-Aware Allocation
  & $78.52$ & $88.87$ & $85.24$ & $93.98$
  & $86.65\,(+2.95)$ & $1.018{\times}$ \\
  Content-Aware Selection
  & $59.96$ & $78.59$ & $67.74$ & $88.17$
  & $73.61\,(-10.09)$ & $1.000{\times}$ \\
  \textbf{AllocEmbed}
  & $\mathbf{83.87}$ & $\mathbf{90.15}$ & $\mathbf{89.20}$ & $\mathbf{94.36}$
  & $\mathbf{89.39\,(+5.69)}$ & $0.977{\times}$ \\
  \bottomrule
\end{tabular*}%
}
\caption{Complete LongRet results corresponding to Figure~\ref{fig:main-longret-2b}. Base embeds 8 uniformly sampled frames. The remaining methods start from 24 candidate frames, with Content-Aware Selection embedding 8. Values are Hit@1 (\%), and Avg. is the mean over four subsets. Parentheses show absolute changes from Base. Cost is normalized to Base ($1.000{\times}$). Bold marks the best budget-matched result in each column. $^\dagger$ Full Resolution processes all 24 frames at the Base per-frame resolution and is not budget matched.}
\label{tab:main-longret-2b}
\end{table*}

\end{document}